# HOLISTIC PARAMETERIC RECONSTRUCTION OF BUILDING MODELS FROM POINT CLOUDS

Zhixin Li [1], Wenyuan Zhang [2, 1], Jie Shan [1]

[1] Lyles School of Civil Engineering, Purdue University, Indiana, USA - (li2887, jshan)@purdue.edu
[2] National Research Center of Cultural Industries, Central China Normal University, Wuhan, China – zhangwy@mail.ccnu.edu

**Commission II, WG II/6**

**KEY WORDS:** Primitives, Building Modelling, Point Cloud, Semantic Segmentation, Deep Neural Network, CityGML

**ABSTRACT:**

Building models are conventionally reconstructed by building roof points planar segmentation and then using a topology graph to group the planes together. Roof edges and vertices are then mathematically represented by intersecting segmented planes. Technically, such solution is based on sequential local fitting, i.e., the entire data of one building are not simultaneously participating in determining the building model. As a consequence, the solution is lack of topological integrity and geometric rigor. Fundamentally different from this traditional approach, we propose a holistic parametric reconstruction method which means taking into consideration the entire point clouds of one building simultaneously. In our work, building models are reconstructed from predefined parametric (roof) primitives. We first use a well-designed deep neural network to segment and identify primitives in the given building point clouds. A holistic optimization strategy is then introduced to simultaneously determine the parameters of a segmented primitive. In the last step, the optimal parameters are used to generate a watertight building model in CityGML format. The airborne LiDAR dataset RoofN3D with predefined roof types is used for our test. It is shown that PointNet++ applied to the entire dataset can achieve an accuracy of 83% for primitive classification. For a subset of 910 buildings in RoofN3D, the holistic approach is then used to determine the parameters of primitives and reconstruct the buildings. The achieved overall quality of reconstruction is 0.08 meters for point-surface-distance or 0.7 times RMSE of the input LiDAR points. The study demonstrates the efficiency and capability of the proposed approach and its potential to handle large scale urban point clouds.

## 1. INTRODUCTION

Highly accurate and semantically rich three-dimensional (3D) building models become inevitable for various applications such as urban planning, urban change detection, 3D navigation, emergency response, and heritage conservation (Gao et al., 2018). Over the past few decades, many research efforts have been conducted to 3D building modelling at different levels of detail (LoD) from multi-source remotely sensed data. However, developing a fully automated and efficient building modelling technique with high accuracy and rich semantics remains a challenging task (Jung et al., 2017).

Since the emerging of laser scanning and oblique photography technologies, there has been remarkable progress in the field of 3D point cloud generation, processing and applications, which accelerates the research and development in 3D building modelling (Rottensteiner et al., 2014). Conventionally, building models are reconstructed by separately fitting roof planes in the point clouds and then using a topology graph to group these planes. In this type of approach, roof edges and vertices are mathematically represented by intersecting segmented planes. Recent research presented some plane-based methods that firstly segment the point clouds into primitive planes and then intersect them into polygonal meshes in 3D space (Bauchet and Lafarge 2019). Technically, such solutions are based on sequential local fitting, i.e., at one time only part of the entire data of a building participates in determining the building model. As a consequence, the solution is lack of topological integrity and geometric rigor. Furthermore, the fully automatic process of the traditional reconstructing approaches remains challenging due to limited point density, sensor noise, missing data, outliers, and scene complexity (Cao et al., 2017; Macher et al., 2017; Zhu et al., 2017).

More recently, deep learning has increasingly gained attention in various applications. For instance, 3D neural networks have been extensively explored for 3D object detection and reconstruction (Zhi et al., 2018), 3D semantic segmentation (Engelmann et al., 2017; Graham et al., 2017; Tchapmi et al., 2017), 3D classification of point clouds (Özdemir et al., 2019; Uy et al., 2019) and 3D object pose estimation (Qi et al., 2019).

Fundamentally different from the traditional approach, we propose a holistic parametric building reconstruction method based on deep neural networks and (3D) primitives. Based on a set of predefined parametric roof primitives in 3D space, we first use a sophisticated deep neural network to segment and identify roof primitives in the building-only point clouds. A holistic optimization strategy is then introduced to simultaneously determine the parameters for a segmented building. Subsequently, a watertight building model can be uniquely reconstructed that best fit the point cloud for 3D modelling and representation. Our contributions can be summarized as: 1). Propose an automatic end-to-end building reconstruction pipeline for building point clouds; 2). Present a holistic parametric estimation method; and 3). Introduce an effective and robust solution of semantic building modelling from point clouds.

The rest of the paper is organized as follows. Section 2 describes related work in primitive fitting, point cloud semantic segmentation and reconstruction. Section 3 introduces the methodology for parametric reconstruction (i.e., determine the values of these parameters). Section 4 tests the proposed algorithm on the RoofN3D dataset (Wichmann et al. 2018) with three predefined primitives. The work exhibits promising results. A quantitative assessment is conducted, followed by a comparative analysis on the results from different roof structures. Section 5 expresses the conclusions drawn from the experimental results.

## 2. RELATED WORKS

### 2.1 Primitive Fitting

The idea of decomposing a complex model into a set of simple geometric primitives for object recognition originates from the concept of recognition-by-components proposed by Biederman (1987). Recognition of primitive types and primitive fitting are key issues and challenging tasks for point cloud segmentation or shape detection. Schnabel et al. (2007) developed an automatic and efficient random sample consensus (RANSAC) based framework for detecting planes, spheres, cylinders, cones and tori in unorganized point clouds. However, under-segmentation and false detection of primitive types may occur since RANSAC-based approach only looks for local cues and its estimation of the primitive parameters is sensitive to noise in position and normal (vector) of the sample points (Isack and Boykov, 2012). Furthermore, the performance of RANSAC-based methods relies on careful and laborious per-input parameter tuning. Le and Duan (2017) proposed a primitive-based 3D segmentation framework for mechanical CAD models. A dimension reduction method was employed to transform the detection of 3D primitives into the classical 2D problems such as circle and line detection in images. Li and Feng (2019) introduced geometric primitive segmentation for convolutional neural network (CNN) and proposed a framework for multi-model 3D primitive fitting based on simulated point clouds. It performed superior than RANSAC-based methods on noisy range images of cluttered scenes. In contrast to traditional parsing or detecting shapes from point cloud data, Li et al. (2019) introduced an end-to-end neural network named supervised primitive fitting network (SPFN) to detect multiple primitives of different types with accurate parameters. This approach supports the prediction of plane, sphere, cylinder, and cones at different scales, and does not require any user intervention. Li et al. (2019) presented a primitive-based 3D building modelling approach by synthesizing the training data. PointNet was adopted to classify building primitives, while coherent point drift (CPD) was applied to align the predicted primitives into the target 3D point clouds.

### 2.2 Semantic Segmentation

Semantic segmentation or classification of point clouds is a well-known problem in computational geometry and computer vision. It has been extensively researched over the past decades. Due to the recent advancements of deep neural networks (DNN) in image semantic segmentation, DNN has been extensively used or extended for 3D semantic segmentation (Qi et al., 2016). To handle irregular point clouds, 3D DNN has been further proposed, including PointNet (Qi et al. 2017), PointNet++ (Qi et al. 2017), VoxelNet (Zhou and Tuzel 2018), VoteNet (Qi et al. 2019). These methods showed remarkable performance in semantic segmentation, classification, 3D object detection of point clouds. Tchapmi et al. (2017) presented an end-to-end framework to obtain 3D point-level segmentation that combines the advantages of neural network, trilinear interpolation and fully connected conditional random fields.

### 2.3 Semantic Modelling

Building models with semantic information, such as CityGML and Industry Foundation Classes (IFC) provide more valuable application potentials than traditional geometric model, which leads to a new research topic about building semantic modelling in recent years. Semantic reconstruction of different building roof types is a crucial task for 3D building modelling. Most of the previous approaches rely on human intervention beyond the selection of processing parameters, which is tedious, time-consuming, as well as the most expensive part of the workflow. In order to reduce labour-intensive processes, many efforts have been made to develop automatic methods over recent years. Zheng et al. (2017) presented a hybrid method for automatic reconstruction of complex building roof structures, in which data-driven approach was used to detect step edges. Roof types were determined by using LiDAR and high-resolution orthophotos. Finally, plane fitting was employed to reconstruct parametric models and generate semantic building models at LoD2. Jayaraj and Ramiya (2018) used commercial software and open source packages to generate 3D building models in CityGML format from airborne LiDAR point cloud. However, most of these approaches rely on multi-source data. Automation and accurate reconstruction still pose great challenges to the existing algorithms.

## 3. METHODOLOGY

As shown in Figure 1, our workflow consists of the following three conceptual steps. The input LiDAR point clouds are first semantically segmented by PointNet++. Then an optimization method with fixed initial values and predefined cost function, consists of point-surface-distance (PSD) and 2D Intersection over Union (IoU), is used to determine the parameters for each individual building with specific primitive type. Finally, semantic building models in CityGML format are reconstructed based these estimated primitive parameters.

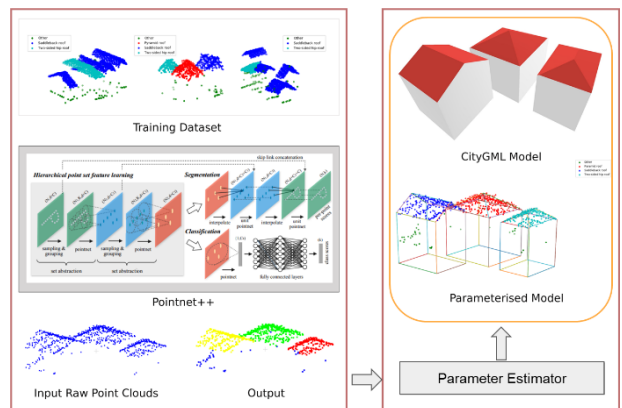

Figure 1. Workflow of parametric reconstruction of building models.

### 3.1 Learning-based Point Cloud Segmentation

Identify roof type of point cloud is the basis of parametric reconstruction of a building. For recognizing various roof types of point clouds, we choose to use a learning-based method to automatically segment input point clouds into different parts with proper roof type labels. To achieve this goal, the hierarchical deep network, PointNet++ (Qi et al. 2017) is integrated into our pipeline. This high-performance network has four sets of abstraction (SA) layers for subsampling the input point clouds and two feature propagation/up-sampling (FP) layers for up-sampling intermediate point features. The output of this step is segmented building points with identified roof types.

A library of basic primitives is firstly defined. At this time, the library consists of several roof types defined in semantic-rich OGC CityGML standard (Gröger et al. 2012), including flat roof, gable roof, hip roof, half-hip roof, shed roof, mono pitch roof, pyramid roof, and mansard roof. These geometric-topological primitives are represented in a local coordinate system by several parameters, such as width, length, height, etc. Additionally, all primitives are placed in the form of axis-aligned elements. Figure 2 shows three standard primitives (pyramid, gable and hip) to be considered in this paper.

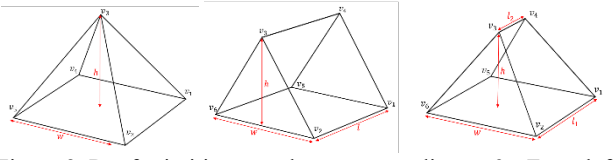

Figure 2. Roof primitive samples corresponding to $\theta_p$. From left to right: pyramid, gable, hip primitives

### 3.2 Parameter Estimation

After labelling roof type of a point cloud from DNN, we start to estimate the corresponding primitive parameters for best fitting given building clouds. In this stage, parameters are separated into three groups: global translation parameters ($T_G$), primitive parameters ($\theta_p$) and local rigid transformation ($\theta_0$), where $\theta_p$ and $\theta_0$ are to be optimized. For the convenience and efficiency of computation, we firstly translate building points $P$ from a world coordinate system to a local coordinate system whose origin is at the centroid of the building points; hence we have the $T_G$.

Next, we establish the relation between the roof primitive parameters and its vertices. This relation can be described as a function $S = f(\theta_p)$ where $S$ represents surfaces, by calculating vertices of each roof surface and determining their connectivity. Take gable primitive as an example, its six (6) vertices can be expressed by $\theta_p = \{w, l, h\}$, as labelled in Figure 3. The two roof planes are constructed with $S_1 = \{v_4, v_3, v_2, v_1\}$ and $S_2 = \{v_3, v_4, v_5, v_6\}$, as shown in Figure 3. To start the optimization, $\theta_p$ is initially estimated by using the minimum bounding box of the input point clouds and the identified primitive type.

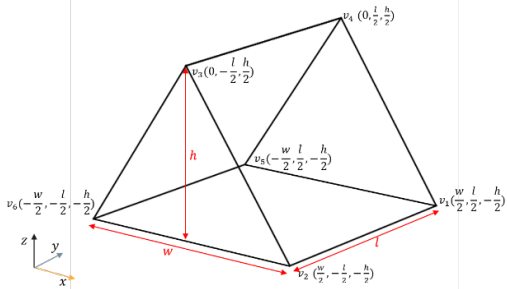

Figure 3. Coordinates for vertices of a gable roof

Then we use $\theta_0$, including local rigid transform matrix based on Euler rotation angle $\kappa$ around $Z$ axis and local translation parameters $T_L$ to transform the primitive surfaces into the correct position during optimization. The reason we only use single angle rotation around Z axis is because most buildings have only one degree of freedom, making them "perpendicular" to the ground. As such, for surface $S_j$, its centroid $[\bar{x}_j\ \bar{y}_j\ \bar{z}_j]^T$ and their centralized vertices $\bar{A}_j$ can be determined through the roof primitive function $f_j(\theta_p, \theta_0)$. Then the coefficient set of $S_j$: $C_j = [a_j\ b_j\ c_j\ d_j]^T$ can be solved through singular value decomposition (SVD) with respect to $\bar{A}_j$ as follows

$$\bar{A}_j = f_j(\theta_p, \theta_0) \quad (1)$$
$$\bar{U}_j \bar{\Sigma}_j \bar{V}_j^T = \bar{A}_j \quad (2)$$

$$[a_j\ b_j\ c_j]^T = \bar{U}_j(:,3) \quad (3)$$
$$d_j = -[a_j\ b_j\ c_j] \cdot [\bar{x}_j\ \bar{y}_j\ \bar{z}_j]^T \quad (4)$$
$$C_j = [a_j\ b_j\ c_j\ d_j]^T \quad (5)$$

where $[a_j\ b_j\ c_j]$ is the normal of $S_j$, given by the third column of $U$.

Finally, we optimize the primitive parameters, local rotation and translation by minimizing the overall cost function $J$ with respect to the PSD and IoU, shown as below

$$J = J_1 + \beta J_2 \quad (6)$$

where

$$J_1 = \frac{1}{N}\sum_{i=1}^{N}\min(g(P_i, C_j) \mid j=1,2,\ldots,m) \quad (7)$$

$$J_2 = 1 - \frac{IntersectArea(S_B, B)}{UnionArea(S_B, B)} \quad (8)$$

In Equation 6, $J$ consists of $J_1$ (mean PSD), $J_2$ (0~1, relative missing area represented by IoU) and $\beta$ (coefficient to balance $J_1$ and $J_2$). $\beta$ is an empirical coefficient based on raw LiDAR RMSE (~0.1 meter), which can make $J_1$ and $J_2$ (~0.01) in the same scale. In Equation 7, $g$ is a function to calculate distance from a point to a plane, $P_i$ is $i^{th}$ point in the input points, $m$ is the number of surfaces in the primitive and $N$ is the total number of points. In Equation 8, $S_B$ is the projected horizontal boundary of a roof primitive and $B$ is the 2D $\alpha$-shape boundary of the input points, and IoU is served as a necessary boundary constrain for primitive parameter optimization. Once the cost function $J$ is formed, the L-BFGS-B method (Zhu et al., 1997) is used to determine $\theta_p$ and $\theta_0$. During optimization, $\theta_p$ and $\theta_0$ will be updated and hence Equation 1-8 are recalculated to form $J$. After minimization of the cost function $J$, roof parameters $\theta_0$ and $\theta_p$, along with building primitive type and global translation $T_G$, will be grouped in a JSON file for the next step.

### 3.3 Parametric Reconstruction

Once we get the estimated parameters for each building, we move to the last stage of building reconstruction. First of all, we calculate the world coordinates of roof vertices by using global translation parameters and optimal local primitive parameters, as well as the labelled roof types. Take the reconstruction of gable roof as an example, the world coordinates of roof vertices are inferred through rotating and translating the primitive to correct pose and position. The whole rigid transformation of vertices is expressed as follows

$$V = R_\kappa V_\theta + T_L + T_G \quad (9)$$

where $V_\theta$ is the local coordinates of vertices with respect to estimated primitive parameter $\theta_p$, $R_\kappa$ is the rotation matrix with respect to optimal angle $\kappa$ in local parameter $\theta_0$, $T_L$ represents the optimal values of local translation, while $T_G$ is the given global translation.

After we get the accurate vertex coordinates of roof structure, different semantic surfaces defined in CityGML LoD2 building model are sequentially reconstructed by using these vertices and the underlying topology. Furthermore, all surfaces are generated with boundary representation (BRep). For instance, two roof surfaces in Figure 3 are represented by $S_{r1} = \{v_4, v_3, v_2, v_1, v_4\}$ and $S_{r2} = \{v_3, v_4, v_5, v_6, v_3\}$ separately to form oriented and closed planar surfaces. The ground elevation with respect to each roof vertex is calculated by combining the digital elevation model (DEM) and its footprint coordinates in the ground. The minimum value $z_{min}$ of all the calculated ground elevations is taken as the base elevation for a building. Thus, the four vertices of horizontal ground are $V_g = \{(v_i(x), v_i(y), z_{min}) | i = 1, 2, 5, 6\}$, and the ground planar surface is created and represented by $S_g = \{v_{g1}, v_{g5}, v_{g6}, v_{g2}, v_{g1}\}$. Four façades are subsequently modelled according to the roof vertices and ground vertices in the same vertical plane, i.e., $\{v_1, v_2, v_{g2}, v_{g1}, v_1\}$, $\{v_2, v_3, v_6, v_{g6}, v_{g2}, v_2\}$, etc. To achieve the desired effect of 3D visualization, all the coordinates of vertices in a surface are specified in counter clockwise order and used to form a <gml:posList> element. To this end, all exterior boundaries of a building are generated into piecewise planar surfaces with semantic information, such as RoofSurface, WallSurface, and GroundSurface. After hierarchically organized into different tree elements according to CityGML building schema and encoding standard, these semantic surfaces are finally form a watertight LoD2 building model with the specific element <bldg:Building>. To represent the volume of the building geometry, a closed LOD2 solid geometry (lod2Solid) is also created within the above building element by referencing to the generated boundary surfaces. Buildings with hip or pyramid roof could be reconstructed in a similar way as gable roof. The semantic buildings models with three different roof structures are shown in Figure 4. Figure 4 (a) is the input three point clouds, red point are roof points, while blue ones are ground or wall points. Figure 4 (b) illustrates the reconstructed CityGML building models.

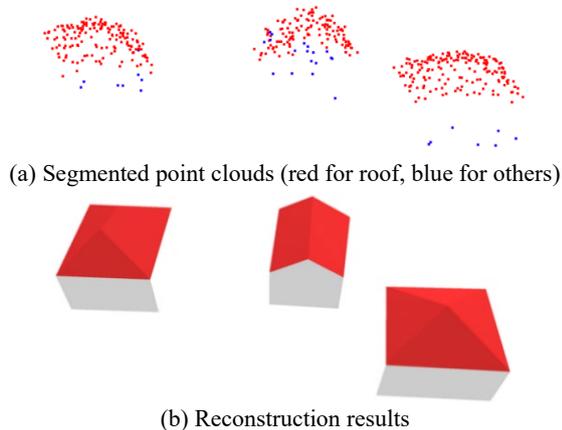

(a) Segmented point clouds (red for roof, blue for others)

(b) Reconstruction results

Figure 4. Reconstructed CityGML models with different roof primitives (hip, gable, pyramid)

In order to integrate building models with terrain, the Terrain Intersection Curve (TIC) defined in CityGML is also considered during the modelling. Thus, a lod2MultiCurve geometry is generated and applied to each building by connecting all of the calculated ground points with their elevation. In addition to the geometric representation of a building, some significant attributes related to a building are created and appended to the building element as well, such as the bounding box (envelope), the name of the spatial reference system (srsName), the measured height of the building (measured Height). Ultimately, a coherent geometric-semantical CityGML LOD2 building model is produced by outputting these elements to a GML file.

## 4. EXPERIMENTS

### 4.1 Dataset

Our testing data is the recently released building dataset RoofN3D. This dataset is a 3D building point clouds dataset for training 3D DNN and building reconstruction. It covers a large area in New York City (NYC) and includes rich semantic information. Provided by the U.S. Geological Survey (USGS), the point clouds were collected from airborne LiDAR from 08/2013 to 04/2014 and cover an area of 1,009.66 sq. km. The average density of the point clouds is about 4.72 points/m2. The spatial reference of point cloud is NAD83(2011)/UTM zone 18N with unit in meters. The whole dataset consists of 118,074 single buildings and is pre-labelled three building types, i.e., pyramid, gable, and hip. As shown in our workflow, for the first DNN step, we used 80% (94,459) buildings as training dataset and the rest 20% (23,615) as testing dataset. For the second estimation step, we use a subset of 910 single buildings (including 81 pyramids, 630 gables, and 199 hips) located in NYC Queens residential area to evaluate the estimator performance. Additionally, corresponding DEM with sub-meter resolution is also included to reconstruct the bottom surface for each building.

### 4.2 Results

Table 1 shows the performance of segmentation and primitive type classification using whole RoofN3D dataset with PointNet++. As shown in Table 1, we achieve an overall 47.85% semantic segmentation IoU and 83.02% roof type classification accuracy on the testing dataset. The classification accuracy demonstrates that most roof structures of the building points can be recognized correctly under the powerful DNN model.

| # Testing buildings | 23,615 |
|---|---|
| # Roof types | 3 |
| Semantic Segmentation IoU | 47.85% |
| Classification Accuracy | 83.02% |

Table 1. Segmentation and classification result of the RoofN3D dataset from PointNet++

A subset of RoofN3D Dataset is then selected to test the quality of building modelling. It consists of 910 buildings with three different roof types. 910 CityGML LoD2 building models are automatically generated by using the proposed parametric reconstruction approach. Part of the reconstructed result is shown in Figure 5. All the roof surfaces are rendered with red colour. As we can see, most models coincide with building footprints, which indicates these models are reconstructed in the right geographic positions and right orientation.

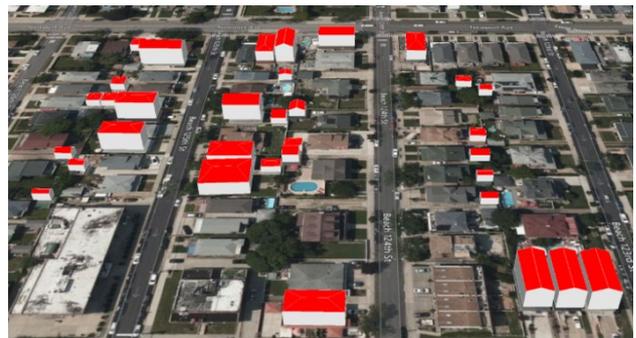

Figure 5. A sample area of the reconstructed CityGML building models in NYC

### 4.3 Quality of the reconstructed models

The quality assessments of reconstruction result are performed in two ways. First, the mean and standard deviation of the PSD from the point cloud to the corresponding reconstructed roof are calculated to validate the roof geometric accuracy. Second, the 2D IoU between bottom surfaces of the reconstructed building model and the corresponding ground truth building footprint is adopted to evaluate the horizontal accuracy.

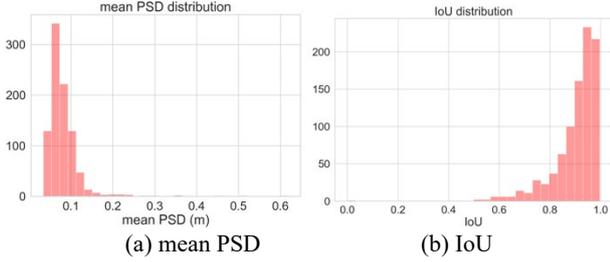

(a) mean PSD     (b) IoU

Figure 6. Error distribution of the reconstructed 910 buildings

The distributions of PSD and IoU metrics are shown in Figure 6, while the statistical results for each roof type are summarized in Table 2.

|  | Pyramid | Gable | Hip | Overall |
|---|---|---|---|---|
| #Buildings | 81 | 630 | 199 | 910 |
| Mean PSD (m) | 0.0630 | 0.0809 | 0.0863 | 0.0805 |
| Std. of PSD(m) | 0.0503 | 0.0692 | 0.0701 | 0.0677 |
| Mean IoU (%) | 90.12 | 88.66 | 95.17 | 90.22 |

Table 2. Quality evaluation of reconstructed models in terms of primitive types

On one hand, the distance between reconstructed roof surfaces and their point clouds is smaller than 0.1m in most cases. The overall PSD is 0.0805m, which mostly indicates the shape (geometry) accuracy of the reconstructed building models. On the other hand, the 2D IoU's for most reconstructed models are higher than 80%, and the overall IoU is 90.22%, which mostly demonstrates the accuracy of building boundary. With respect to three kinds of roof types, the reconstruction of hip roofs achieves the best boundary accuracy, while the pyramid roofs receives a shape accuracy better than the others. The difficulty of finding optimal parameters in search space grows exponentially with the number of parameters. With respectively two (2) and one (1) additional parameters to be optimized, hip and gable roofs achieve a shape accuracy slightly poorer than the pyramid. Furthermore, there are quite a few asymmetric gable buildings in the test dataset, which may cause their slightly lower boundary accuracy comparing to the pyramid and hip buildings.

### 4.4 Stability of the Approach

To test the stability of our optimization approach, we implement a pipeline to generate simulated points with permutated errors and estimate corresponding parameters. We start this pipeline with a set of parameters estimated from building roof points from the RoofN3D dataset. We select 20 buildings for each building type and generate 30 trials of independent and identically distributed random points for each building.

|  |  | Pyramid | Gable | Hip | Overall |
|---|---|---|---|---|---|
| # buildings |  | 20 | 20 | 20 | 60 |
| Dimension (w, l, h,…) | RMSE(m) | 0.120 | 0.217 | 0.462 | 0.266 |
|  | RMSE(%) | 4.37 | 5.55 | 5.90 | 5.273 |
|  | STD (m) | 0.071 | 0.136 | 0.297 | 0.168 |
| Translation | RMSE(m) | 0.020 | 0.046 | 0.030 | 0.032 |
|  | STD (m) | 0.018 | 0.042 | 0.030 | 0.030 |

Table 3. Performance of the optimization algorithm with simulated points

Based on the RoofN3D's metadata, we add random noise to the simulated points with RMSE as 0.12 meters and the noise is distributed as: 90% within 0~1 RMSE, 9% within 1~2 RMSE and 1% within 2~3 RMSE, which is similar to the real data. The optimization algorithm is then used to estimate the parameters of the simulated points. The results are then compared with the "true" parameters. Table 3 summarizes the root mean square error, its percentage relative to the roof dimension, and the standard deviation of the differences between estimated parameters and the "true" parameters.

### 4.5 Discussions

As we carefully check each building model and its quantitative evaluation indexes, there exists several less accurate results with either a higher mean PSD or a lower IoU. For instance, a specific building labelled with gable roof got a high mean PSD (mean PSD=0.0687m) and low IoU (IoU=69.73%). The comparison of the reconstructed roof and its referenced boundary model is shown in Figure 7. It shows that the distribution of the input point cloud is not symmetric, whereas the predefined gable primitive is a fully symmetric structure. As such, the result of primitive fitting is erroneous in this case.

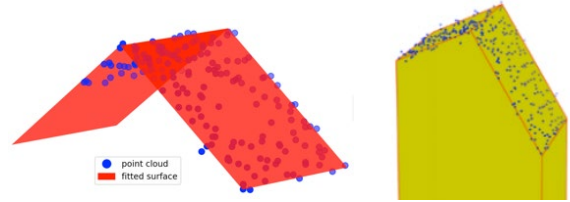

Figure 7. Primitive-fitted roof surface (left) and reference boundary model from RoofN3D (right).

In contrast, building roofs may be occluded by small surrounding trees or other objects, which leads to an incomplete point cloud. Under such circumstance, the proposed method is able to reconstruct a complete and symmetric roof surfaces. The solution is robust to incomplete point cloud data, a superior property over traditional data-driven approaches (e.g. RANSAC, region growing). In addition, Figure 8 shows comparison of results between RANSAC plane detection and the proposed holistic approach. Figure 8 left indicates two undetected planes (red boxes), though there are substantial sampling points on a hip roof. However, the primitive based holistic fitting could lead to correct roof structure.

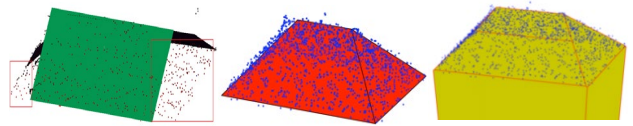

Figure 8. Comparison between RANSAC plane detection (left), reconstructed roof (middle) and ground truth (right).

Although the RoofN3D dataset provides boundary models of what, the ground height for all the models are 0, which is not able to be used as ground truth for building height. Thus, the vertical accuracy with respect to ground elevation is difficult to evaluate by using the RMSE between reconstructed models and ground truth.

Furthermore, RoofN3D only offers three categories of roof structure. As such, our assessment was only conducted to limited roof structures. However, the proposed parametric reconstruction algorithm is flexible and can be easily extended to model buildings with complex roof structure. This is mostly because of a number of typical primitives defined in the library and deep learning based semantic segmentation of point cloud.

Furthermore, since most complex buildings can be recognized and decomposed into several simple primitives by the DNN model, each segmented building part can be reconstructed separately through this optimization process for primitive fitting. At the end, all building parts can be composed into the complete model with topological and semantic rules.

## 5. CONCLUSION

This paper presented a novel holistic parametric reconstruction method. All points of a building are taken into account simultaneously to determine the building parameters and its symmetry. To best fit a point cloud with corresponding roof primitive, both roof surfaces and their projected horizontal boundary are considered through a carefully designed objective function. The point-surface-distance helps define the shape of the roof primitives, while the projected horizontal boundary is used to form a necessary boundary condition. Experiments demonstrated this can overcome certain limitations of previous sequential single or simple plane fitting-based approaches. The primitive based reconstruction results have several advantages such as symmetry, regularization, and compact representation. Moreover, the new development is robust to noise, outliers, and missing data.

Experimental results with over 900 buildings showed that the proposed method can accurately and effectively generate semantic building models with several different roof structures. The approach is stable for the tested three types of buildings, with 5.27% dimension difference and can achieve 0.08 meter point-to-surface distance, or 0.7 times the RMSE of the input LiDAR points. Furthermore, the entire reconstruction process is fully automatic and is implemented using Python. As such, it can be used to model large-scale scenes with highly dense urban areas.

The generated models are represented according to CityGML building encoding, which offers the advantages of spatio-semantic coherence, geometrical-topological coherence, as well as level of detail. These semantic models can be applied to a wide range of fields.

However, the reconstruction quality is limited by the accuracy of deep learning based point cloud segmentation, since the rooftop semantic segmentation results are important input for the subsequent parametric estimation and building reconstruction. Additionally, reconstruction quality with respect to other roof categories and complicated roof structures needs to be furtherly assessed by exploring more point cloud data with ground truth. Future work could be on exploring more effective DNN and optimization techniques, and extending this work to reconstruction of compound buildings.